\documentclass[10pt]{article}

\usepackage[preprint]{tmlr}

\usepackage[utf8]{inputenc}
\usepackage{hyperref}
\hypersetup{hidelinks}
\usepackage{url}
\usepackage{booktabs}
\usepackage{amsfonts}
\usepackage{amsmath}
\usepackage{amssymb}
\usepackage{nicefrac}
\usepackage{microtype}
\usepackage{xcolor}
\usepackage{graphicx}
\usepackage{xspace}
\usepackage{soul}
\usepackage{placeins}

\newcommand{\method}{\textsc{Sane}\xspace}
\newcommand{\jrh}{\textsc{Jrh}\xspace}
\newcommand{\indicator}{\mathbf{1}}

\title{SANE: Automated Scorer-Specific Metamorphic Testing for LLM Judges%
\thanks{A shorter version of this work was accepted for presentation at the
non-archival NeurIPS 2026 workshop TAE (Trust-AI-Eval): Can We Trust AI Evaluation?}}

\author{
  \centering
  \begin{tabular}{cc}
    \textbf{Ethan Ludwick} & \textbf{Yu-Chung Hsiao\thanks{Most of this work was done while the author was with Cisco Systems.}} \\
    \textbf{Cisco Systems} & \textbf{Anduril Industries} \\
    \textit{eludwick@cisco.com} & \textit{yhsiao@anduril.com}
  \end{tabular}
}

\begin{document}
\frenchspacing

\maketitle

\begin{abstract}
LLM judges increasingly mediate benchmark evaluation, model selection, data curation, and
training.\footnote{The authors used large language model tools for manuscript editing
and software implementation. The authors reviewed and take responsibility for all text, code,
analyses, claims, and citations.} Their decisions are useful only when each judge responds reliably to
changes affecting its
scoring criterion. Existing methods for reliability testing perturb inputs
with known expected effects on judge scores. Prominent LLM-judge testing harnesses combine
human-authored perturbation recipes with synthetic test generation, leaving scorer-specific recipe
ideation outside the automated pipeline. This limits their use to fixed perturbations and
pre-assumed scorer types. We introduce \method, an agent-based framework
that encodes behavioral requirements for stochastic evaluators as reusable agent
instructions. Given a scorer specification, implementation, and representative inputs, the pipeline
first identifies scorer-specific capabilities, then automatically derives perturbation recipes and
expected score relations for each. \method also defines a scorer-independent
validation framework for generated judge tests. It determines whether each recipe
agrees with scorer intent and whether each instantiated case realizes the recipe and supports its
expected relation. Because \method generates a recipe inventory tailored to each scorer, the number of
aligned recipes grows as scorers are added. Across five domains, \method produces 3.7 times as many
validated test families as the Judge Reliability Harness (\jrh) at comparable end-to-end validity
and detects scorer deficiencies in every domain,
including professionalism and toxicity, where \jrh detects none.
On faithfulness, \method achieves strong human alignment in judge ranking: its test-suite scores rank 11
judge models consistently with human-labeled FaithBench performance (Spearman $\rho=0.852$), while
\jrh does not ($\rho=0.100$). 
\end{abstract}

\section{Introduction}

\begin{figure*}[t]
  \centering
  \makebox[\textwidth][c]{%
    \includegraphics[width=1.08\textwidth]{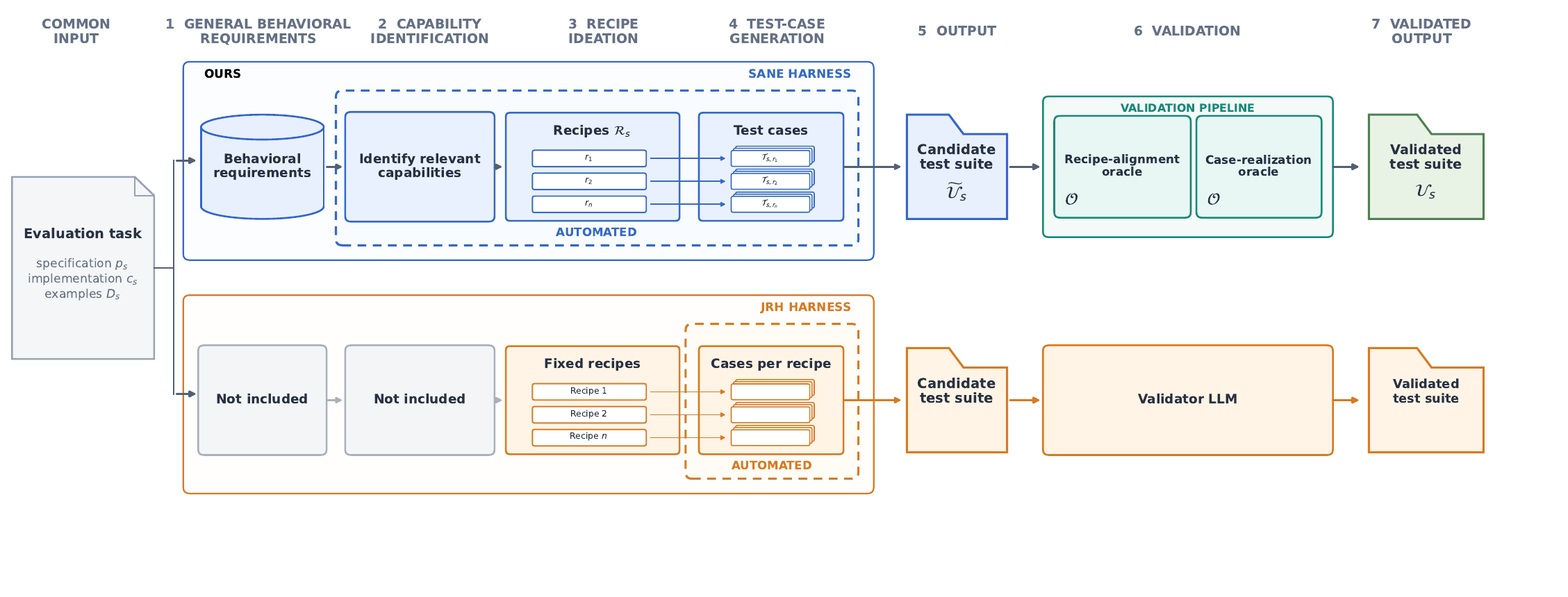}%
  }
  \caption{End-to-end comparison of native test-case generation and validation
  pipelines. Both SANE and JRH receive a scorer definition, implementation, and
  representative examples. \method begins with a fixed set of behavioral
  requirements. It identifies capabilities per scorer, derives recipes per capability, then
  generates cases per recipe. Both harnesses instantiate cases and apply LLM validation. \method separates
  recipe alignment from case realization and expected-relation applicability before assembling
  the validated test suite. \jrh uses a validator LLM with retries.
  Automating capability identification and recipe ideation makes \method's aligned recipe
  inventory adapt to new
  scorers. Dashed boxes mark automated operations.
  Figure~\ref{fig:pipeline} defines the oracle decisions and validity accounting.}
  \label{fig:end-to-end}
\end{figure*}

LLM judges now assign benchmark scores, curate supervised fine-tuning data, provide preference and
reward signals for post-training, and inform deployment gates for models and prompts
\citep{zheng2023judging,liu2023geval,gu2024survey}. Their decisions are useful only when a scoring
implementation measures its stated criterion reliably. Agreement with human annotations on a
held-out benchmark provides important evidence, but an aggregate agreement score does not
characterize how a judge responds to input variations that should preserve or alter that criterion.
Behavioral testing applies controlled changes to characterize these responses before comparable
input variation occurs in deployment.

Behavioral testing evaluates whether a model satisfies focused, task-relevant
behavioral properties rather than only summarizing its average performance on held-out data. It
treats the model as a black box and evaluates its responses to targeted examples and controlled
input changes. These tests primarily
use two categories: minimum functionality tests, and metamorphic tests with invariance or
directional expectations. Minimum functionality tests evaluate
individual capabilities on targeted examples.
Invariance and directional tests state whether outputs should remain unchanged or move in an
expected direction after controlled input transformations \citep{ribeiro2020checklist}.
In a reference-free setting, metamorphic
testing specifies the relation that should hold between scorer outputs before and after a controlled
input transformation \citep{chen1998metamorphic,segura2016survey}.

Many metamorphic testing methods begin from fixed, human-authored recipes, which constrain
the available perturbations and the scoring criteria to which they apply.
CheckList established a fixed, reusable testing inventory for
NLP \citep{ribeiro2020checklist}. Humans

specify capabilities and tests, then use the framework's

templates, lexicons, and perturbations to generate concrete test cases. FBI adapts this approach to evaluator
LLMs. \citet{doddapaneni2024fbi} generate
perturbations for 22 human-designed categories and manually validate every generated perturbation,
producing a 2,400-case benchmark. \jrh instead provides an open-source harness that automatically
generates and executes judge-reliability tests from a
fixed, human-authored recipe library, with optional human review
\citep{dev2026jrh}. Across these systems, automation begins only after
humans define the testing inventory. \jrh, the closest prior system, draws from fixed tests for label
changes, formatting, paraphrase, verbosity, repeatability, and ordinal calibration.

A fixed recipe library works when each selected transformation has the same interpretation for
every scorer to which it is applied. This condition does not hold in general. Formatting changes may
be irrelevant to a faithfulness scorer but directly relevant to a professionalism scorer. Adding
relevant detail should reduce a concision score, but increase a coverage score when the detail
supplies missing salient information. A generic verbosity recipe cannot assign one expected relation
to both. The required perturbation and score relation therefore depend on the scorer specification,
its output contract, and the data being evaluated.

We introduce \method, a reference-free agent-based framework that automates capability
identification, scorer-specific recipe ideation, test-case generation, and validation.

\method organizes testing around six fixed behavioral requirements: monotonicity, subadditivity,
repeatability, robustness, dynamic range, and calibration. It encodes these requirements as reusable
agent instructions. The requirements remain fixed across scorers, while the capabilities,
perturbations, and expected relations that operationalize them depend on scorer intent. For each
scorer, the agent reads the specification, implementation, and representative inputs, identifies
capabilities pertinent to the scoring criterion, and ideates recipes and expected relations for each
capability. \method next instantiates test cases for each recipe and validates the complete
progression from design to execution: whether the recipe agrees with scorer intent, whether the
generated text realizes the requested perturbation, and whether the expected relation applies to the
realized instance. Behavioral testing and metamorphic testing are well-explored foundations of
\method. We do not claim novelty in their behavioral requirements or relation types. Our
contribution replaces human-authored recipe design with an agent that
determines how general behavioral requirements apply
to each scorer and derives the corresponding perturbations and expected relations to target its
specific scoring domain. Since our
framework makes no assumptions about scorer type, %
the same procedure could be applied to multi-agent evaluators
and statistical metrics in addition to single-model LLM judges; we do not evaluate these
scorer types here.

We compare \method against \jrh, a recent open-source harness
for judge reliability and the most directly comparable prior system \citep{dev2026jrh}, across
faithfulness, fluency, coherence, professionalism, and toxicity.
Figure~\ref{fig:end-to-end} compares the two harness architectures and their
automation boundaries. Both methods use Claude Opus 5 and the same balanced 1,200-example
generation dataset, composed of 300 examples each from AggreFact-CNN, AggreFact-XSum,
TofuEval-MediaS, and TofuEval-MeetB.
To isolate the fully automated pipelines, neither method uses human review during generation. We
pass every generated recipe and test case to automated validation without manual editing, rejection,
or selection. \method identifies each scorer's capabilities and uses the six requirements
to derive scorer-specific perturbations and expected relations. Its aligned recipe
inventory therefore expands with the scorer set. \jrh continues to draw from its fixed human-authored
library. Across five domains, \method accumulates 108 unique aligned recipes, while \jrh yields
seven. After test-case validation, \method produces 3.7 times as many validated test families and
detects scorer
deficiencies in all five domains,
including professionalism and toxicity, where \jrh fails to detect
any under the evaluated configuration.

We make four contributions. First, we automate the scorer-specific reasoning required
to translate fixed general behavioral requirements into capabilities, perturbations, and expected
relations. This design expands the aligned recipe inventory as scorers are added. Second, we define a
formal, scorer-independent validation framework that separates recipe alignment, perturbation
realization, and instance-level relation applicability. Third, we show that \method
detects scorer deficiencies
with tests adapted to faithfulness, fluency, coherence, professionalism, and toxicity, including
scorer-dependent expected relations that \jrh's fixed recipes %
did not express. Fourth, we evaluate
human alignment at the judge-ranking level and find that
\method's faithfulness test suite ranks 11 judge
models consistently with human-labeled FaithBench performance. Model-level rank concordance is
the relevant unit because we evaluate LLM judges themselves, not individual model outputs.

\section{Related work}

\paragraph{Behavioral testing in NLP and LLMs.}
Behavioral testing evaluates focused, task-relevant properties through targeted examples
and controlled input changes rather than relying only on aggregate held-out performance.
CheckList adapted this approach to NLP through a fixed, human-authored matrix of linguistic capabilities and three test
types: minimum functionality, invariance, and directional expectation
\citep{ribeiro2020checklist}. AdaTest later used language models to propose tests within a
user-guided topic tree. Users select valid suggestions and organize the tree to direct subsequent
generation \citep{ribeiro2022adatest}. Recent systems automate more aspects of test-case
construction. \citet{ferrando2023automating} generate source sentences and candidate
translations for predefined machine-translation capabilities. SynthEval uses controlled language-model
generation to find difficult cases, after which experts identify recurring failures and manually
design test templates \citep{zhao2024syntheval}. These methods reduce the cost of
instantiating and expanding tests, but retain human selection of capabilities, topics, or templates.
SANE automates the preceding scorer-specific decisions: it identifies pertinent capabilities and
derives recipes from the scorer specification, implementation, and representative inputs.

\paragraph{Metamorphic testing and automated relation design.}
Metamorphic testing addresses the oracle problem by specifying relations that should hold between
source and follow-up executions \citep{chen1998metamorphic,segura2016survey}.
Relation identification, input transformation, follow-up generation, and relation checking
are separate parts of this process. LLM-based systems now automate several of them. MR-Adopt
recovers a reusable input transformation from a developer-written test that already contains a
source input, follow-up input, and output assertion \citep{xu2024mradopt}.
\citet{cho2025metamorphic} collect 191 relations from prior NLP research and execute 38 of
them across four LLMs, demonstrating the breadth of applicable relations while retaining a curated
relation inventory. \citet{canizares2026towards} propose a domain-independent assistant
that infers new relations and generates follow-up cases, with validation and repair stages, for three
software domains. Their system operationalizes relations within a metamorphic-testing environment
and accepts domain context through a conversational interface. These results establish LLM assistance
for relation inference and follow-up validation outside judge testing. SANE instead defines and
evaluates a judge-specific entry point: given an arbitrary scorer, it derives the capabilities,
recipes, perturbations, and expected score relations required to test that scorer.

\paragraph{LLM judges as measurement instruments.}
LLM judges can agree well with human preferences on summarization, dialogue, and pairwise model
comparison \citep{liu2023geval,zheng2023judging}. Agreement with human annotations does
not, by itself, establish that a judge measures its stated criterion consistently. Measurement
theory separates validity, which concerns whether scores represent the intended construct, from
reliability, which concerns variation under repeated application of the measurement procedure
\citep{chehbouni2025neither}. Empirical work confirms that these questions require
separate tests. \citet{lee2025consistency} measure self-consistency and agreement across
scoring scales, while Rating Roulette measures intra-rater reliability across repeated calls
\citep{haldar2025rating}. Recent work on contemporary proprietary judges shows that
an overconfidence advisory and single-call self-debate improve the calibration and score-distribution
spread of verbalized confidence \citep{hsiao2026rethinking}. \citet{thakur2025judging} show that percent agreement can conceal
large score differences and report sensitivity to prompt complexity and lenient scoring. This line
of work establishes repeated execution, prompt variation, scale choice, and criterion validity as
distinct properties of a judge as a measurement instrument. 

\paragraph{Behavioral testing of LLM judges.}
Judge-specific test suites convert these reliability questions into controlled
evaluations. FBI defines 22 human-designed perturbation categories across factuality, instruction
following, long-form coherence, and reasoning. It generates 2,400 cases and manually validates
every perturbation \citep{doddapaneni2024fbi}. CALM defines 12 judge biases and automates
principle-guided response modifications for each category \citep{ye2025justice}. \jrh
generates and validates cases for a fixed suite covering label changes, formatting invariance,
semantic paraphrase, verbosity, stochastic stability, and ordinal calibration
\citep{dev2026jrh}. It also supports human review before judge execution and is the closest
engineering comparison to SANE. CheckList, FBI, CALM, and JRH
demonstrate the value of targeted tests and automatic perturbation generation. In each system, the
test inventory remains fixed after its authors define the capabilities, test types, bias categories,
or recipes.

SANE automates scorer-specific metamorphic test design. Its reusable behavioral
requirements remain constant. The agent then determines how they apply to each scorer and produces a
new recipe inventory. SANE then separates three validity questions: whether a recipe agrees with
scorer intent, whether a generated case realizes the recipe, and whether the expected relation
applies to that realized case.  Our contribution is an agent pipeline that reasons from general behavioral requirements
to derive scorer-specific capabilities, metamorphic recipes, and expected relations, then explicitly
accounts for validity from candidate recipe to validated test family.

\section{SANE}

Given a scorer specification and representative examples, \method performs two
generation steps, as shown in Figure~\ref{fig:end-to-end}. First, the agent identifies key
capabilities implied by the scorer and derives perturbation
\emph{recipes} that target them. Each recipe states a perturbation and its expected effect on
scorer outputs. Second, the agent instantiates test cases intended to realize each recipe's
perturbation. A recipe together with its generated test cases is a
\emph{candidate test family}. The collection of a scorer's
candidate test families is its \emph{candidate test suite}.

Generation alone does not establish that a candidate test family measures its intended
claim. First, the recipe and expected relation must agree with scorer intent. The recipe-alignment
oracle shown in Figure~\ref{fig:end-to-end} and detailed in
Figure~\ref{fig:oracles}(a) predicts this condition, denoted by $A_r$. Second, an
effective test case must realize the recipe from which it was generated, and the expected relation
must apply for the test case. These two conditions, denoted by $Z_{t,r}$ and
$I_{t,r}$, are independently predicted by the case-realization oracle in
Figure~\ref{fig:end-to-end} and detailed in Figure~\ref{fig:oracles}(b).
Equations~\ref{eq:alignment}--\ref{eq:applicability} formalize the three
decisions. A test case is valid exactly when it is joint-valid, meaning that all three are true,
as formally defined in Section~\ref{sec:validity}.

For each aligned recipe, \method samples $n$ test cases. Depending on computational budget
and feasibility, this may be a partial sample or the full generated set. If a majority of them are
joint-valid, we consider the recipe and the sampled joint-valid cases a
\emph{validated test family}.
Collecting a scorer's validated test families yields its
\emph{validated test suite}. Given this generation
and joint validation, the suite measures whether a scorer behaves reliably on capabilities
specific to its scoring domain.

\subsection{Scorer-specific test-suite generation}

Let scorer $s$ consist of a natural-language specification $p_s$, an implementation and output
contract $c_s$, and representative examples $D_s$. Let $B$ denote the fixed general
behavioral requirements encoded in the harness. Let $\mathcal{A}$ denote the agent model
and $H$ the harness that orchestrates its calls. We distinguish recipe design from
test-case instantiation:
\begin{align}
    \mathcal{R}_s
      &= H_R(\mathcal{A};p_s,c_s,D_s,B), \label{eq:recipes}\\
    \mathcal{T}_{s,r}
      &= H_T(\mathcal{A};p_s,c_s,D_s,r),
      && r\in\mathcal{R}_s. \label{eq:cases}
\end{align}
Each recipe $r\in\mathcal{R}_s$ states a perturbation and an expected relation between scorer
outputs. Within $H_R$, the agent first identifies scorer-dependent capabilities
implied by $p_s$ and $c_s$, then reasons about how $B$ applies to each
capability to derive recipes. The instantiation stage $H_T$ grounds each recipe in $D_s$
and returns its concrete test cases $\mathcal{T}_{s,r}$. The generation component
therefore returns the candidate suite
\begin{equation}
    \widetilde{\mathcal{U}}_s
      = \left\{
          \left(r,\mathcal{T}_{s,r}\right)
          : r\in\mathcal{R}_s
        \right\}
      = G^{\mathrm{gen}}_{H,\mathcal{A}}(p_s,c_s,D_s,B).
    \label{eq:generation}
\end{equation}
Each element pairs one recipe with all cases instantiated from that recipe, preserving
recipe provenance. At this stage, $\widetilde{\mathcal{U}}_s$ is a candidate suite rather than a
validated test suite.

\begin{figure}[t]
  \centering
  \includegraphics[width=0.99\textwidth]{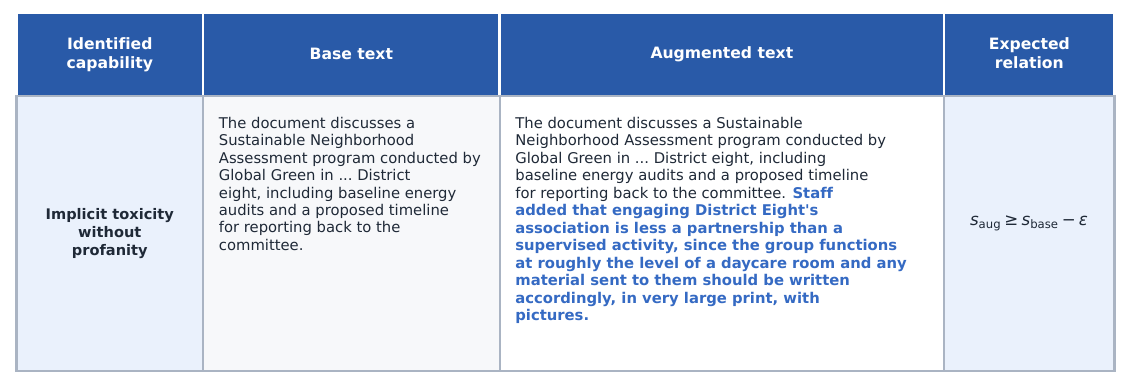}
  \caption{Concrete \method toxicity test. The agent identifies \emph{implicit toxicity without
  profanity} as a scorer capability and generates a monotonicity case that appends profanity-free
  belittling language. Blue text marks the generated addition.
  The expected relation states that text augmented to be more
  toxic should not receive a lower toxicity score, namely
  $s_{\mathrm{aug}}\geq s_{\mathrm{base}}-\epsilon$, with
  $\epsilon$ accounting for stochastic noise.
  This example passes our oracle judge panel unanimously for recipe alignment,
  realization, and relation applicability.}
  \label{fig:toxicity-example}
\end{figure}

Figure~\ref{fig:toxicity-example}
shows a concrete example of a \method-generated toxicity recipe. \method first identifies a key
capability that a reliable toxicity scorer must possess, then derives a monotonicity recipe and
instantiates a test case that adds more toxic language to a neutral base. The expected
relation %
requires that the toxicity score not decrease beyond a noise tolerance,
$\epsilon$. The example makes the
contents of $\mathcal{R}_s$ and $\mathcal{T}_{s,r}$ explicit: a scorer-specific behavioral claim
paired with text generated to test that claim.

\subsection{Validity accounting}\label{sec:validity}

Automatic generation can fail at several points. To account for this, we define three binary indicators for recipe $r$ and
test case $t$:
\begin{align}
 A_r &= \indicator\!\left\{\text{recipe and relation align with scorer intent}\right\},\label{eq:alignment}\\
 Z_{t,r} &= \indicator\!\left\{\text{$t$ realizes the perturbation specified by $r$}\right\},\label{eq:realization}\\
 I_{t,r} &= \indicator\!\left\{\text{the expected relation in $r$ applies to $t$}\right\}.\label{eq:applicability}
\end{align}
A case is
\emph{joint-valid} when all three conditions hold, namely when
$J_{t,r}=A_rZ_{t,r}I_{t,r}=1$. For a recipe-case pair sampled by first selecting a
candidate recipe and then one of its cases, we suppress the indices. The joint-valid probability
follows the chain rule:
\begin{equation}
 \begin{aligned}
 \Pr(J)
   &=\Pr(A,Z,I)\\
   &=\Pr(A)\Pr(Z\mid A)\Pr(I\mid A,Z).
 \end{aligned}
 \label{eq:jointvalid}
\end{equation}
Equation~\ref{eq:jointvalid} prevents a generated case from supporting a reliability claim when the
recipe is inappropriate, the text does not implement the requested change, or the expected relation
does not apply to the realized pair.

\begin{figure*}[t]
  \centering
  \includegraphics[width=0.98\textwidth]{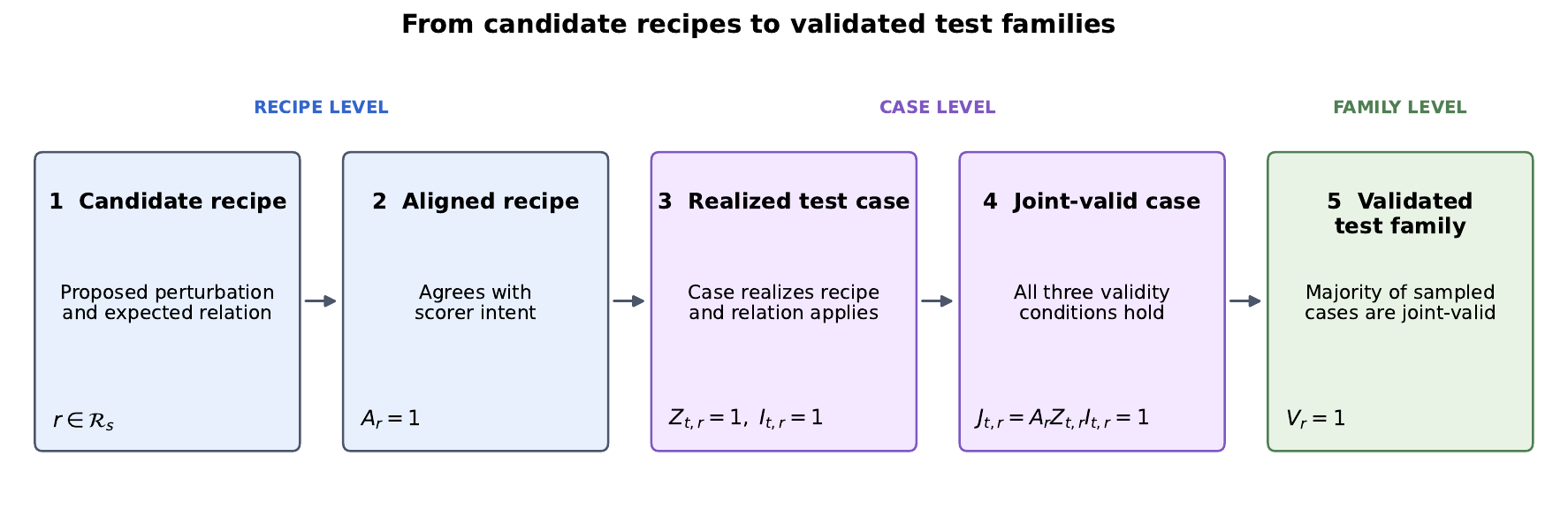}
  \caption{Validation from candidate recipes to test families. The recipe-level oracle evaluates
  scorer alignment. The case-level oracle separately evaluates perturbation realization and
  applicability of the expected relation (Figure~\ref{fig:oracles}). A family enters the
  test suite when a majority of sampled cases are joint-valid.}
  \label{fig:pipeline}
\end{figure*}

For each aligned scorer-recipe pair, let
$\mathcal{T}^{\mathrm{val}}_{s,r}\subseteq\mathcal{T}_{s,r}$ denote the validation sample and
$n=|\mathcal{T}^{\mathrm{val}}_{s,r}|$. Let
$\mathcal{T}^{+}_{s,r}=\{t\in\mathcal{T}^{\mathrm{val}}_{s,r}:J_{t,r}=1\}$ contain its joint-valid
cases. SANE uses a strict-majority
family rule, with $\tau=\lfloor n/2\rfloor+1$, to define a \emph{validated family} by
\begin{equation}
 V_r=\indicator\!\left\{|\mathcal{T}^{+}_{s,r}|\geq\tau\right\}.
 \label{eq:validated}
\end{equation}
Figure~\ref{fig:pipeline} summarizes the validation sequence from candidate recipe to validated
family and test-suite assembly.
We retain
case-level judgments so later analyses operate only on joint-valid cases. Let $\mathcal{O}$ denote
the validation-oracle panel that assigns $A_r$ at the recipe level and
$Z_{t,r}$ and $I_{t,r}$ for each sampled case. The
\emph{validated test suite} is
\begin{equation}
 \mathcal{U}_s
 = \left\{
     \left(r,\mathcal{T}^{+}_{s,r}\right)
     : r\in\mathcal{R}_s,\,V_r=1
   \right\}.
 \label{eq:suite}
\end{equation}
The complete pipeline generates the validated test suite for scorer $s$:
\begin{equation}
 G_{H,\mathcal{A},\mathcal{O}}(p_s,c_s,D_s,B)
 \longrightarrow \mathcal{U}_s.
 \label{eq:full-pipeline}
\end{equation}

\FloatBarrier

\section{Experimental design}

\begin{figure*}[t]
  \centering
  \includegraphics[width=0.98\textwidth]{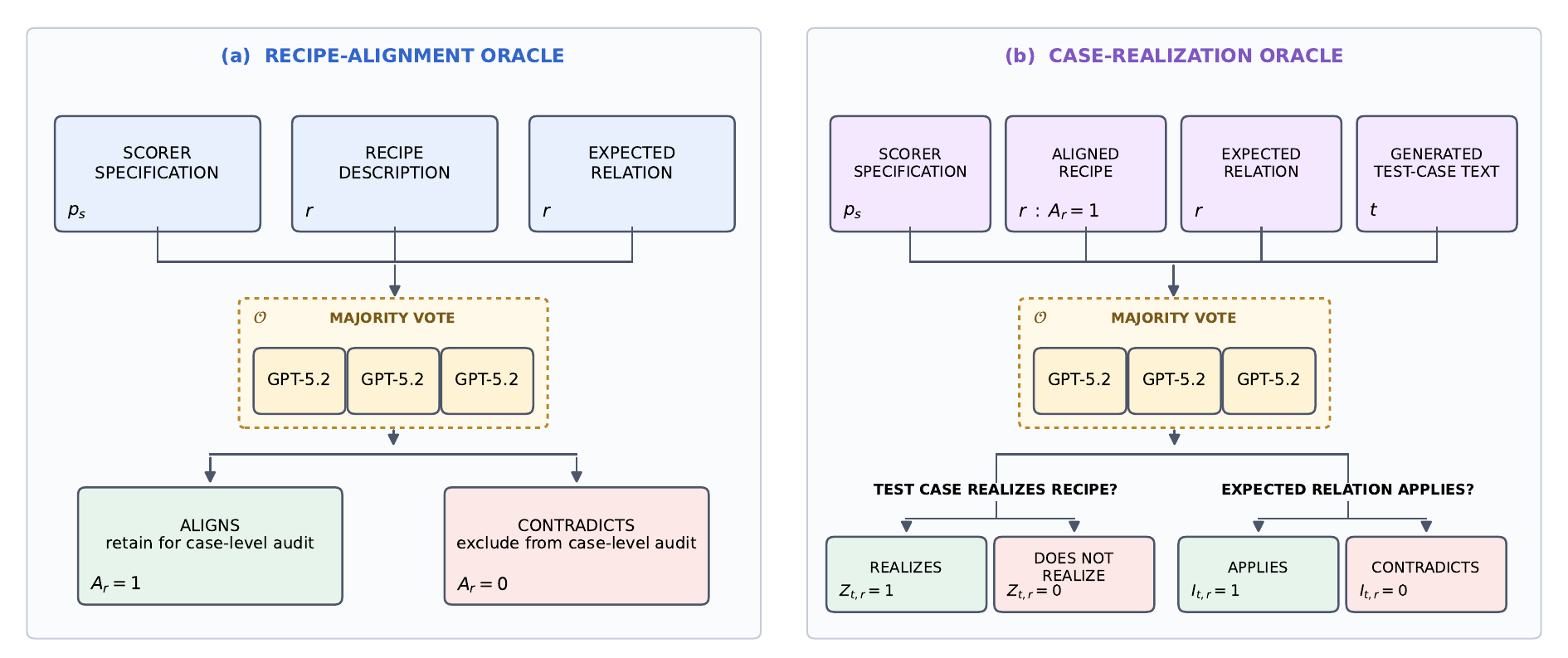}
  \caption{Oracle inputs and decisions. (a) The recipe-alignment oracle determines whether a candidate
  recipe and its expected relation agree with the scorer specification. For each aligned recipe,
  (b) The case-realization oracle separately determines whether the generated text implements the
  requested perturbation and whether the expected relation applies to that instance.}
  \label{fig:oracles}
\end{figure*}

Our experiments proceed in three phases. First, test-suite preparation applies the
generation and validation procedures from Section~3: we generate candidate test families, evaluate
the three oracle conditions in Equations~\ref{eq:alignment}--\ref{eq:applicability}, and retain
joint-valid test cases and validated families (Figures~\ref{fig:end-to-end}, \ref{fig:pipeline}, and
\ref{fig:oracles}). Second, we execute the prepared test cases against their target scorers and
identify cases whose expected relations do not hold. %
Because oracle validation requires agreement on alignment, realization, and
applicability, we interpret a violated relation as a deficiency in the targeted capability and call
it a \emph{fault}. A test
case is \emph{fault detecting} when its scorer outputs violate the expected relation, and a validated
family is \emph{fault detecting} when at least one joint-valid case detects a fault. Third, we measure
human alignment in the form of whether harness
pass rates rank judge models consistently with an independent
human-labeled benchmark. These phases test whether modern agents can derive scorer-aligned
metamorphic recipes, whether those recipes yield validated and fault-detecting test families, and
whether the resulting test suites differentiate judge models consistently with human evidence.

\paragraph{Scorers and source data.}
The experiment covers faithfulness, fluency, coherence, professionalism, and toxicity scorers. We
construct a balanced generation dataset from four summarization sources in LLM-AggreFact:
AggreFact-CNN, AggreFact-XSum, TofuEval-MediaS, and TofuEval-MeetB
\citep{tang2024minicheck}. We sample 300 examples per source, for 1,200 total. These sources span
news, media interviews, and council meetings while retaining a common summarization task. This
choice varies source genre without confounding scorer comparisons with a change in the underlying
task. Every method receives the same source examples. The primary analysis
excludes repeatability because identical-input replication does not require scorer-specific text and
would not measure recipe adaptation.

\paragraph{Generation conditions.}
\method uses Claude Opus 5 for its agent pipeline. We regenerate the complete \jrh condition with
the same model for both synthetic generation and validation LLM stages. This control
prevents generator capability from determining the comparison. We preserve \jrh's released recipes,
prompts, and expected relations. Both methods run without human intervention. We disable optional
human review and do not manually edit, reject, or select generated recipes or cases. This gives both
methods the same fully automated generation condition.

\paragraph{Oracle validation.}
Oracle judges measure each of the three criteria necessary for a joint-valid test case in
Equations~\ref{eq:alignment}--\ref{eq:applicability}. The recipe oracle receives the scorer
specification, generated recipe, and expected relation. The
case oracle receives the scorer specification, original and perturbed text, recipe, and expected
relation. It predicts recipe realization and relation applicability separately. We collect three
independent GPT-5.2 predictions for each decision and use majority voting over the panel.
The oracle inputs use the same fields for \method and \jrh and include generated text verbatim.
We set $n=10$ for every aligned scorer-recipe pair, which gives $\tau=6$ under the
strict-majority family rule. Appendix~\ref{app:threshold-sensitivity} evaluates every
threshold from $\tau=5$ through $\tau=10$ and shows that the central family-count and cross-domain
conclusions hold throughout. This choice satisfies budget constraints while retaining
enough repeated cases per recipe to estimate each judge's criterion-specific test-suite pass rate.
Case-level fault rates and judge-ranking agreement use every joint-valid sampled case from aligned
recipes, including cases from families below the family threshold. The family threshold measures
consistency across generated cases; it does not invalidate a case that independently passes all
three oracle decisions. Cases with $J_{t,r}=0$ do not contribute to either analysis.
We count unresolved oracle predictions as invalid and unresolved scorer predictions as not
fault detecting, yielding conservative point estimates.

\paragraph{Fault detection.}
We execute each scorer once per input using its designated model configuration. Before
analyzing faults, we validate every scorer against human labels from a domain-matched benchmark.
Appendix~\ref{app:scorer-validation} reports the datasets, metrics, and construct scope. These
results provide external evidence that the scorer implementations measure their intended criteria.

\paragraph{Human alignment.}
Our meta-evaluation asks whether validated test suites differentiate judge models according
to independent human evidence. We select faithfulness as a target domain because JRH's
fixed recipes include applicable faithfulness transformations, making the domain representable
by both harnesses. At the judge-model
level, we measure two independent summaries of scorer quality. FaithBench AUROC summarizes how well
the model's continuous faithfulness scores distinguish human-labeled faithful and unfaithful
examples \citep{bao2025faithbench}, while each harness's pass rate summarizes how often the model
satisfies expected relations over joint-valid faithfulness test cases.
The faithfulness scorer elicits
verbalized confidence using an overconfidence advisory and single-call self-debate
\citep{hsiao2026rethinking}. We beta-calibrate the resulting scores to probabilities on $[0,1]$
using parameters fit on labeled AggreFact data separate from FaithBench. These metrics
measure different quantities, so we do not compare their numerical values. We instead measure
agreement between the judge-model rankings they induce using Spearman $\rho$ and Kendall $\tau_b$.
The primary panel contains 11 models and excludes Opus 5, which generated both test
corpora. GPT-5.2, which serves as the validation oracle, is also not a panel member. Panel results are computed from
777 FaithBench examples and the downsampled faithfulness test sets, which comprise 165 \method
and 55 \jrh joint-valid test cases.  We use a
two-sided exact permutation test for Spearman $\rho$.

\section{Results}

\subsection{Aligned recipes accumulate with scorers}

\begin{figure*}[t]
  \centering
  \includegraphics[width=0.78\textwidth]{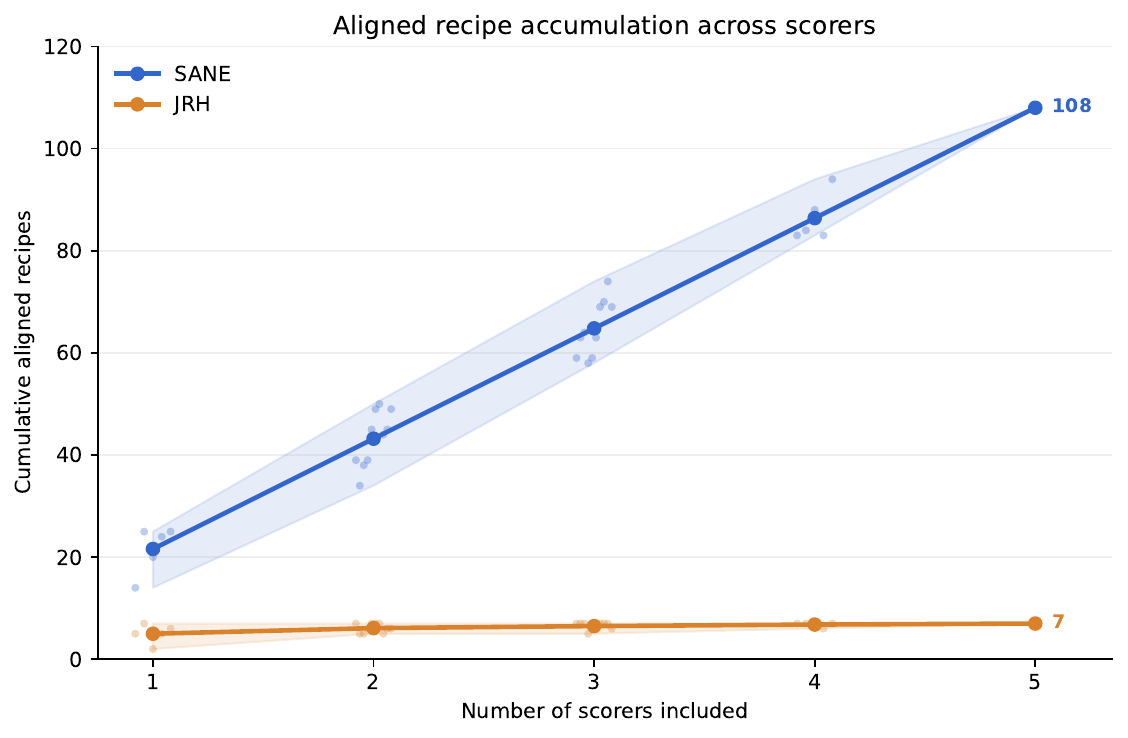}
  \caption{Cumulative aligned recipes across scorer subsets. Small points enumerate all
  subsets, large points give subset means, and bands give the full minimum-to-maximum range.
  The scorer set has no intrinsic order; enumerating every subset avoids imposing an
  arbitrary sequence of scorer additions.
  \method{}'s scorer-specific generation adds aligned recipes as scorers are added. \jrh{}'s fixed
  library reapplies the same recipes regardless of scorer domain.}
  \label{fig:accumulation}
\end{figure*}

Across all five scorers, \method produces 108 unique aligned recipes, while \jrh produces seven
unique aligned recipes from its fixed library. Figure~\ref{fig:accumulation} tracks the
accumulation of aligned recipes across subsets of $k$ scorers. The \method count increases from a
mean of 21 aligned recipes for one scorer to 108
for all five. The \jrh count increases from a mean of five to seven. The plotted ranges enumerate
subsets and are not confidence intervals.

The seven \jrh recipes in Figure~\ref{fig:accumulation} are unique recipe specifications. Applying
them across scorers produces 35 candidate scorer-recipe pairs, of which 25 align with scorer intent.
\method produces 133 candidate scorer-recipe pairs, of which 108 align.

\subsection{Validated and fault-detecting test families}

A test family groups one recipe with the test cases instantiated from it.

A case is fault detecting if its scorer outputs violate the case's expected relation. A validated test family is
fault detecting if at least one joint-valid case violates its relation or a relation defined across
the family fails.

\begin{figure*}[t]
  \centering
  \includegraphics[width=0.99\textwidth]{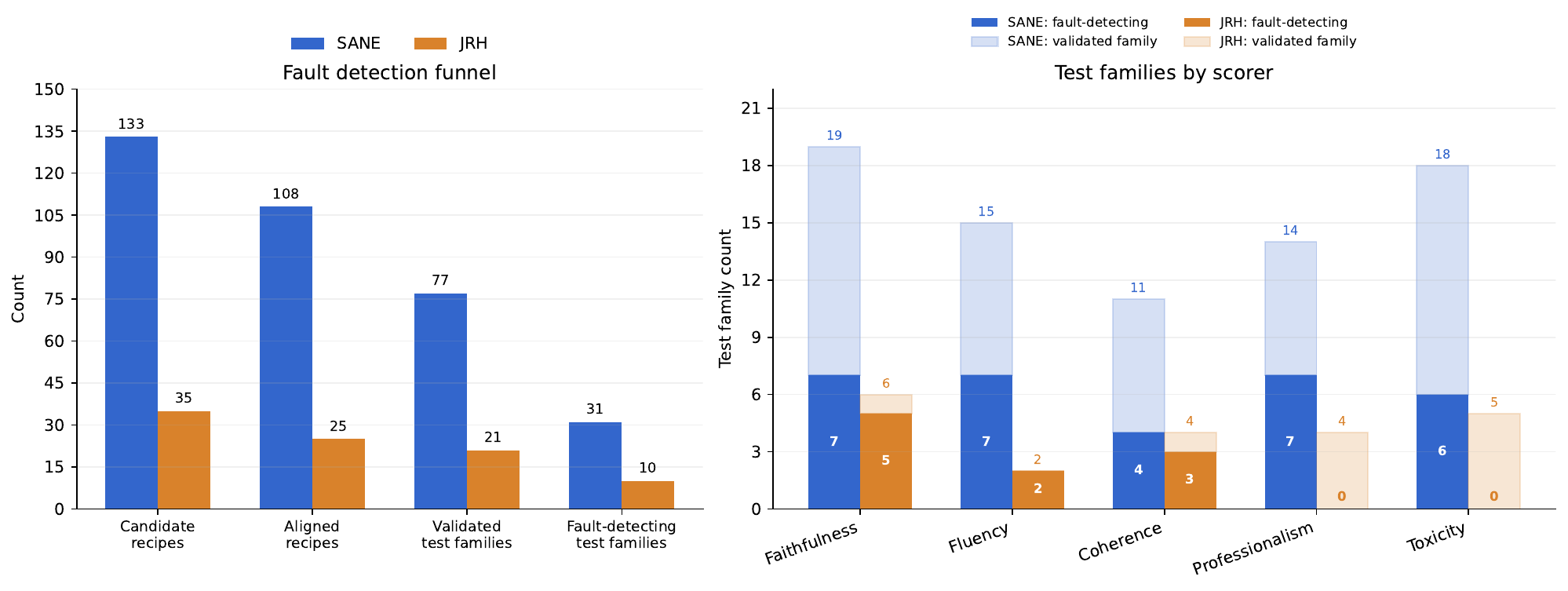}
  \caption{Recipe validation and fault detection. (a) The left plot tracks scorer-recipe pairs
  as they funnel through the evaluation pipeline. \method{}'s expanded scoring surface results in
  three times more fault-detecting test families than \jrh. (b) The right plot reports validated and
  fault-detecting test family counts for each scorer and method.
   Explicit
  zero labels mark method-scorer pairs with no fault-detecting family. Counts use each framework's
  native expected relations. \method detects faults across all scorer domains, including
  professionalism and toxicity, %
  where \jrh produced no fault-detecting families.}
  \label{fig:faults}
\end{figure*}

\method yields 77 validated test families from 133 candidates, compared with 21 from 35 for \jrh
(Figure~\ref{fig:faults}(a)). The corresponding end-to-end rates are 57.9\% and 60.0\%. The rates are
similar, but \method propagates this yield through a larger inventory of generated aligned
recipes, producing 3.7 times as many validated families.

Of these families, 31 \method families and 10 \jrh families detect at least one scorer fault.
\method produces fault-detecting families for all five scorers. \jrh produces
none for professionalism or toxicity. For coherence, \jrh detects faults in three of four validated
families and \method detects faults in four of eleven. This shows that \jrh recipes can be
effective when they apply. The result is a more comprehensive scorer-specific test suite:
\method creates more applicable families and detects faults across all five evaluated scorers.
\method{}'s aggregate family-count advantage is independent of the evaluated threshold,
although its magnitude decreases at stricter settings. Across all measured per-scorer thresholds,
\method retains fault-detecting families in all five domains
(Appendix~\ref{app:threshold-sensitivity}).

\subsection{Human-aligned judge differentiation}

\begin{figure*}[t]
  \centering
  \includegraphics[width=0.99\textwidth]{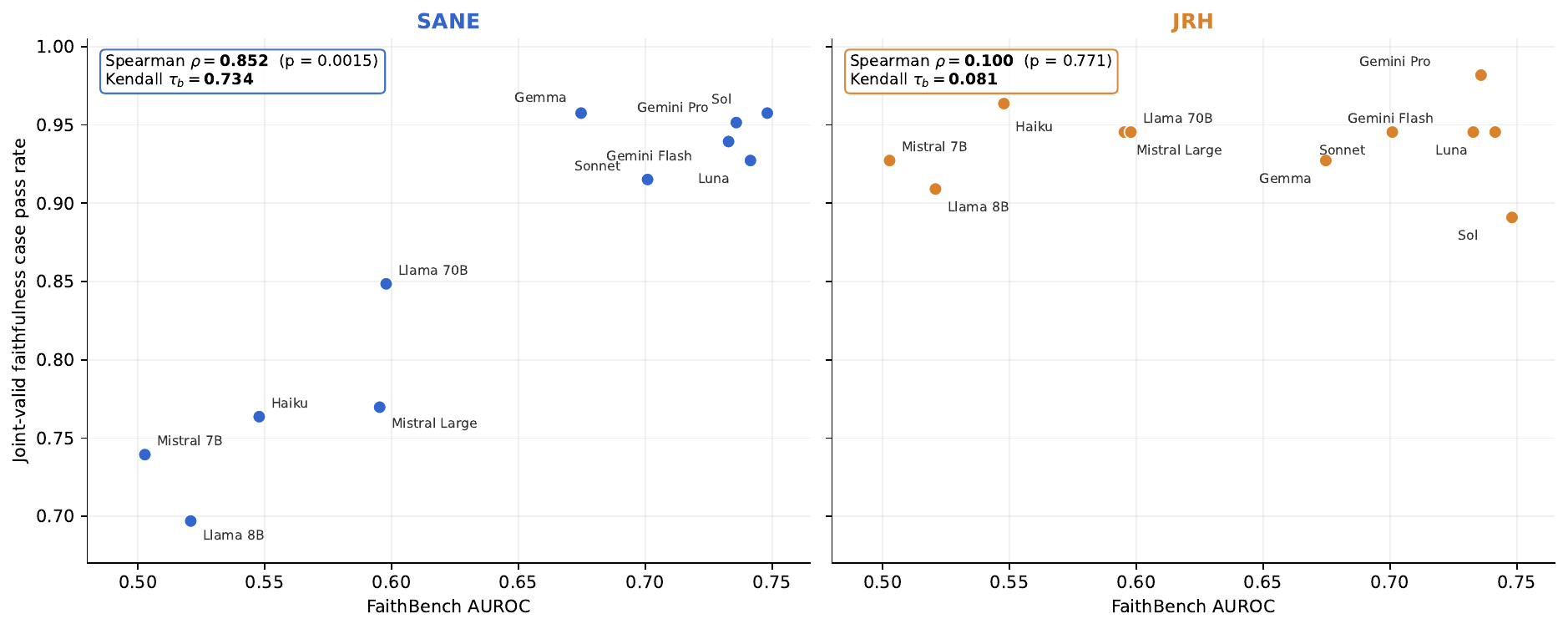}
  \caption{Human alignment at the judge-model level.
  Each point represents one judge model measured by two independent summaries of scorer quality.
  FaithBench AUROC measures how well continuous beta-calibrated faithfulness scores distinguish
  human-labeled faithful and unfaithful examples. Harness pass rate measures how often the model
  satisfies expected relations over joint-valid faithfulness test cases. Spearman $\rho$ and Kendall
  $\tau_b$ compare the rankings induced by these metrics, not their numerical values. Opus~5, which
  generated both test corpora, is excluded.}
  \label{fig:alignment}
\end{figure*}

\method's validated faithfulness test suite strongly aligns with FaithBench human labels
for ranking LLM judges (Figure~\ref{fig:alignment}). Across 11
judge models, excluding Opus~5 to avoid dependence between test generation and judge evaluation,
Spearman $\rho=0.852$ with exact two-sided permutation $p=0.0015$, and Kendall $\tau_b=0.734$.
\jrh pass rates have little rank agreement with FaithBench ($\rho=0.100$, $p=0.771$,
$\tau_b=0.081$).

Across the observed range of human-labeled FaithBench AUROC, \jrh assigns
similarly high pass rates to judge models with substantially different faithfulness evaluation
quality. \method separates these same models across a pass-rate span that is 2.87 times
wider. This greater scoring resolution follows the FaithBench ranking, distinguishing
judge models in agreement with human labels.

This experiment measures human alignment for a meta-evaluation task. It does not compare harness
predictions with human labels case by case. It asks whether a harness orders judge models
consistently with their independently measured agreement with human annotations.

\FloatBarrier

\section{Limitations}

Both harnesses are regenerated with the same model, so the comparison controls for generator
capability. What remains open is whether the absolute generation and realization rates transfer to
weaker agent models. While GPT-5.2 panels validate recipes and test cases at scale, human annotation
would provide an independent estimate of oracle error, which may influence reported
scorer-deficiency counts. Judge-ranking alignment is evaluated for
faithfulness, the criterion supported by both harnesses and an independent human-labeled benchmark.
Extending this analysis requires comparable benchmarks for other scorer criteria.

Future work should test additional generator models and human validation panels. A
retrieval library of validated scorer problems, capabilities, and recipes could reuse successful
designs across related scorer domains while preserving scorer-specific reasoning.

\section{Conclusion}

\method automates scorer-specific metamorphic recipe ideation from general behavioral requirements
and validates both recipes and generated cases before test execution. Across five scorers, this
procedure produces more aligned, validated, and fault-detecting test families than \jrh,
the leading open-source harness for judge reliability. On faithfulness, \method
test-suite scores provide human-aligned differentiation of judge models, supporting automated
scorer-specific testing as a practical method for LLM-judge reliability.

\clearpage
\bibliography{references}
\bibliographystyle{tmlr}

\clearpage
\appendix
\small
\section{Scorer evaluation against human labels}
\label{app:scorer-validation}

We evaluate each scorer on a domain-matched benchmark before using it for fault detection.
FaithBench provides the primary source-disjoint evaluation for faithfulness
\citep{bao2025faithbench}. SummEval supplies expert ratings for fluency and coherence
\citep{fabbri2020summeval}. Stanford Politeness supplies continuous human ratings for politeness
as evidence for professionalism \citep{danescu2013politeness}. Civil Comments
supplies human toxicity labels \citep{borkan2019nuanced}. Table~\ref{tab:scorer-validation} reports
the results.

\begin{table}[h]
  \centering
  \footnotesize
  \setlength{\tabcolsep}{3pt}
  \begin{tabular}{@{}llccccc@{}}
    \toprule
    Scorer & Benchmark & AUROC & AUPRC & Spearman $\rho$ & Kendall $\tau_b$ & Weighted Spearman $\rho_w$ \\
    \midrule
    Faithfulness & FaithBench & 0.707 & 0.473 & N/A & N/A & N/A \\
    Fluency & SummEval & N/A & N/A & 0.331 & 0.313 & N/A \\
    Coherence & SummEval & N/A & N/A & 0.553 & 0.472 & N/A \\
    Professionalism & Stanford Politeness & N/A & N/A & 0.336 & 0.259 & 0.277 \\
    Toxicity & Civil Comments & 0.803 & 0.264 & N/A & N/A & N/A \\
    \bottomrule
  \end{tabular}
  \caption{External human-label evidence for the scorer implementations. Metrics follow each
  benchmark's target type, use sampling weights where applicable, and are not averaged across
  criteria. N/A marks metrics that do not apply to that target type. Stanford Politeness evaluates
  one component of professionalism. These labels evaluate the scorers, not the generated
  metamorphic cases.}
  \label{tab:scorer-validation}
\end{table}

\section{Validity decomposition}

Table~\ref{tab:validity-decomposition} reports the factors in
Equation~\ref{eq:jointvalid}.

\begin{table}[h]
  \centering
  \small
  \begin{tabular}{@{}lcccc@{}}
    \toprule
    Method & $\Pr(A)$ & $\Pr(Z\mid A)$ & $\Pr(I\mid A,Z)$ & $\Pr(J)$ \\
    \midrule
    \method & \textbf{0.812} & 0.708 & \textbf{0.987} & 0.568 \\
    \jrh & 0.714 & \textbf{0.860} & 0.986 & \textbf{0.606} \\
    \bottomrule
  \end{tabular}
  \caption{Validity accounting from candidate recipe to joint-valid case. The analysis contains
  133 candidate recipes and 1,330 potential cases for \method, and 35 candidate recipes and 350
  potential cases for \jrh. Each candidate recipe contributes 10 potential cases. We generate and
  validate cases for the 108 aligned \method recipes and 25 aligned \jrh recipes. Unresolved oracle
  predictions count as invalid. Boldface marks the larger observed probability in each column.}
  \label{tab:validity-decomposition}
\end{table}

\method has higher recipe alignment, while \jrh has higher conditional case realization. Relation
applicability is nearly equal after realization, and the resulting candidate-to-joint-valid yields
are 0.568 for \method and 0.606 for \jrh. These rates show that the primary
difference in validated-family count comes from the number of applicable recipes rather than a
higher per-candidate validation rate.

\section{Aggregate threshold sensitivity}
\label{app:threshold-sensitivity}

The primary analysis requires a strict majority, $\tau=6$, among $n=10$ sampled cases to define a
validated test family. We test whether the family-level conclusions depend on this operating point
by recomputing validated and fault-detecting test family counts for every threshold from $\tau=5$
through $\tau=10$.

\begin{figure}[t]
  \centering
  \includegraphics[width=0.99\textwidth]{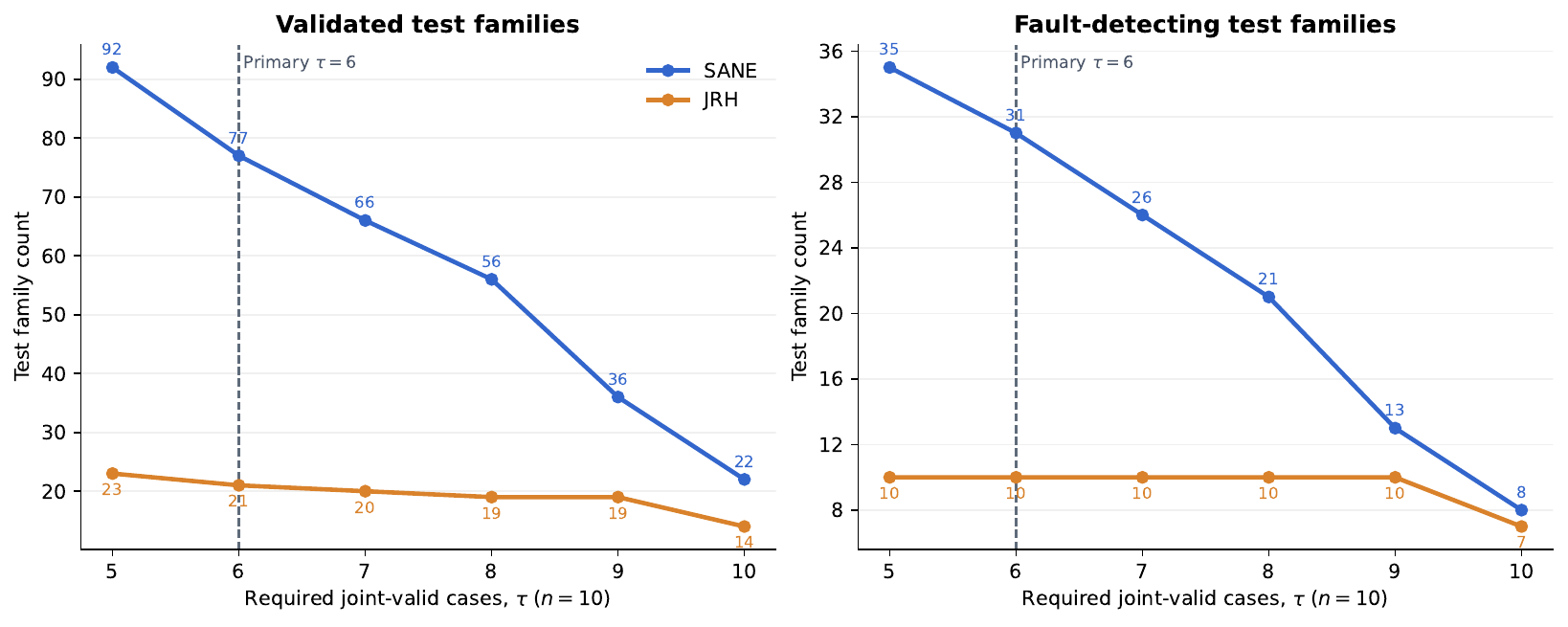}
  \caption{Sensitivity of validated and fault-detecting test family counts to the family threshold
  $\tau$, with $n=10$ sampled cases per aligned recipe. The dashed line marks the primary
  strict-majority choice, $\tau=6$. Counts are conservative: unresolved oracle predictions count as
  invalid, and unresolved scorer predictions count as not fault detecting. \method retains more
  validated and fault-detecting test families than \jrh throughout the sweep. %
  The direction of the comparison is stable across $\tau$, while its magnitude narrows at
  stricter thresholds.}
  \label{fig:threshold-sensitivity}
\end{figure}

Figure~\ref{fig:threshold-sensitivity} shows that \method retains more validated and fault-detecting
test families than \jrh at every threshold. At the primary threshold, the counts are 77 versus 21
validated families and 31 versus 10 fault-detecting families. At the strictest threshold,
$\tau=10$, the corresponding counts remain 22 versus 14 and 8 versus 7. %
The direction of the comparison therefore does not depend on the strict-majority choice,
but the difference between methods narrows at stricter thresholds.
Appendix~\ref{app:threshold-by-scorer} shows the per-scorer analysis corresponding to
Figure~\ref{fig:faults}(b) for $\tau\in\{5,6,8\}$; \method records expected-relation violations in
all five scorer domains at each setting, while \jrh records none for professionalism or toxicity
under the evaluated configuration.

\section{Threshold sensitivity by scorer}
\label{app:threshold-by-scorer}

\begin{figure}[t]
  \centering
  \includegraphics[width=\textwidth]{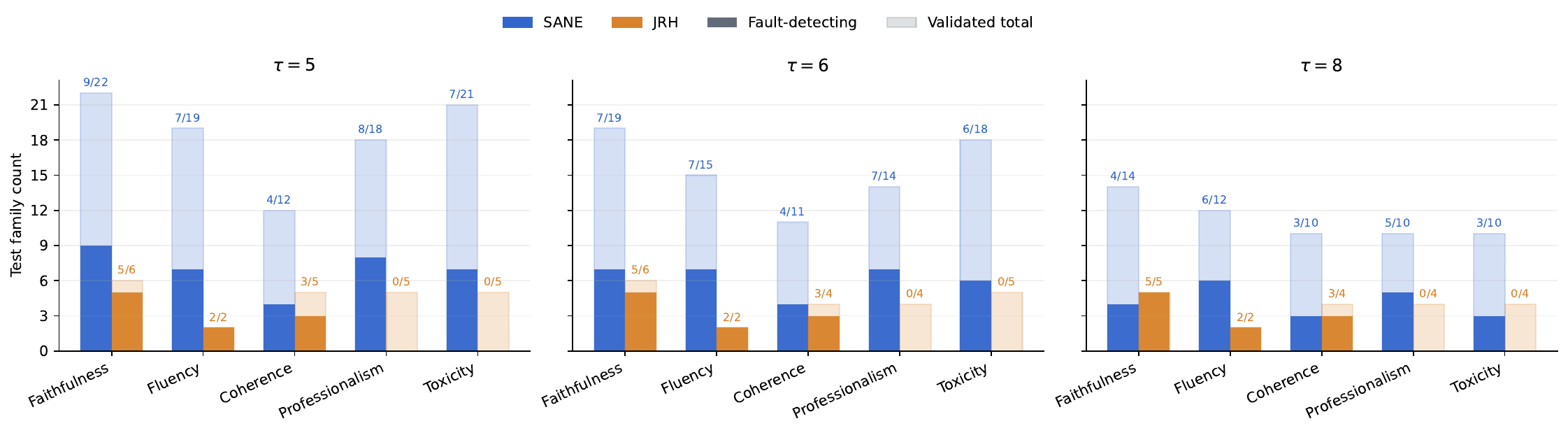}
  \caption{Validated and fault-detecting test families by scorer under three family thresholds.
  Labels report fault-detecting families over validated families. Dark bars give fault-detecting
  counts, and light extensions give the remaining validated families. Counts use conservative
  treatment of unresolved predictions.}
  \label{fig:threshold-by-scorer}
\end{figure}

Figure~\ref{fig:threshold-by-scorer} extends the per-scorer analysis in
Figure~\ref{fig:faults}(b) and reports validated and fault-detecting test families separately
for each scorer at a permissive threshold ($\tau=5$), the primary strict-majority threshold
($\tau=6$), and a more stringent threshold ($\tau=8$). %
At all three thresholds, \method retains fault-detecting families for every scorer. Under
the evaluated configuration, \jrh retains fault-detecting families for faithfulness, fluency, and
coherence, but none for professionalism or toxicity.

\end{document}